\pdfoutput=1

\documentclass[11pt]{article}

\usepackage[final]{coling}

\usepackage{times}
\usepackage{latexsym}

\usepackage[T1]{fontenc}

\usepackage[utf8]{inputenc}

\usepackage{microtype}

\usepackage{inconsolata}

\usepackage{graphicx}

\usepackage{enumitem}
\usepackage{rotating}

\usepackage{booktabs}
\usepackage{multirow}
\usepackage{subcaption}

\usepackage{amsthm}
\usepackage{amsmath}
\usepackage{amssymb}
\usepackage{xspace}
\newcommand{\narrowtextsc}[1]{\textls[-50]{\textsc{#1}}}
\newcommand{\lm}[1]{\texttt{#1}}
\newcommand{\sys}[1]{\narrowtextsc{#1}}
\newcommand{\data}[1]{\textsf{#1}}

\usepackage{xcolor,colortbl}

\definecolor{Gray}{gray}{0.94}
\definecolor{LightCyan}{rgb}{0.88,1,1}
\newcolumntype{a}{>{\columncolor{Gray}}c}
\newcolumntype{o}{>{\columncolor{white}}c}
\usepackage{soul}
\definecolor{celeste}{cmyk}{0.3922, 0.0353, 0, 0.1}
\definecolor{purple}{cmyk}{0.16, 0.28, 0, 0}
\definecolor{brilliantlavender}{cmyk}{0, 0.2235, 0, 0.1}
\definecolor{LightRed}{RGB}{232, 56, 107} 
\definecolor{LightBlue}{RGB}{116, 232, 226}
\definecolor{Tan}{rgb}{0.8203,0.7031,0.5469}
\definecolor{gblue}{RGB}{81,231,195}

\definecolor{greenblue}{RGB}{142, 207,201}

\definecolor{orange}{RGB}{255, 190, 122}

\definecolor{red}{RGB}{250, 127,111}

\definecolor{blue}{RGB}{130, 176, 210}

\title{Cross-Refine: Improving Natural Language Explanation \\ Generation by Learning in Tandem
}

\newcommand{\affilsup}[1]{\rlap{\textsuperscript{\normalfont#1}}}

\author{
    Qianli Wang\affilsup{1,2}
    \qquad 
    Tatiana Anikina\affilsup{1,3}
    \qquad 
    Nils Feldhus\affilsup{1}\\
    \textbf{Simon Ostermann\affilsup{1,3,4}}
    \qquad
    \textbf{Sebastian M\"oller\affilsup{1,2}}
    \qquad
    \textbf{Vera Schmitt\affilsup{1,2}}
    \\
    $^1$German Research Center for Artificial Intelligence (DFKI) \\
    $^2$Technische Universit\"at Berlin \qquad
    $^3$Saarland Informatics Campus\\
    $^4$ Centre for European Research in Trusted AI (CERTAIN) \\
    \texttt{\{firstname.lastname\}@dfki.de}
}

\begin{document}
\maketitle
\begin{abstract}
Natural language explanations (NLEs) are vital for elucidating the reasoning behind large language model (LLM) decisions. Many techniques have been developed to generate NLEs using LLMs. However, like humans, LLMs might not always produce optimal NLEs on first attempt. Inspired by human learning processes, we introduce \sys{Cross-Refine}\footnote{\url{https://github.com/qiaw99/Cross-Refine}}, which employs role modeling by deploying two LLMs as \textit{generator} and \textit{critic}, respectively. The generator outputs a first NLE and then refines this initial explanation using feedback and suggestions provided by the critic. \sys{Cross-Refine} does not require any supervised training data or additional training. We validate \sys{Cross-Refine} across three NLP tasks using three state-of-the-art open-source LLMs through automatic and human evaluation. We select \sys{Self-Refine} \cite{madaan-2023-selfrefine} as the baseline, which only utilizes self-feedback to refine the explanations. Our findings from automatic evaluation and a user study indicate that \sys{Cross-Refine} outperforms \sys{Self-Refine}. Meanwhile, \sys{Cross-Refine} can perform effectively with less powerful LLMs, whereas \sys{Self-Refine} only yields strong results with \lm{ChatGPT}. Additionally, we conduct an ablation study to assess the importance of feedback and suggestions. Both of them play an important role in refining explanations. We further evaluate \sys{Cross-Refine} on a bilingual dataset in English and German.
\end{abstract}


\section{Introduction}

\begin{figure}[ht!]
\centering
\resizebox{\columnwidth}{!}{
\begin{minipage}{\columnwidth}
\includegraphics[width=\columnwidth]{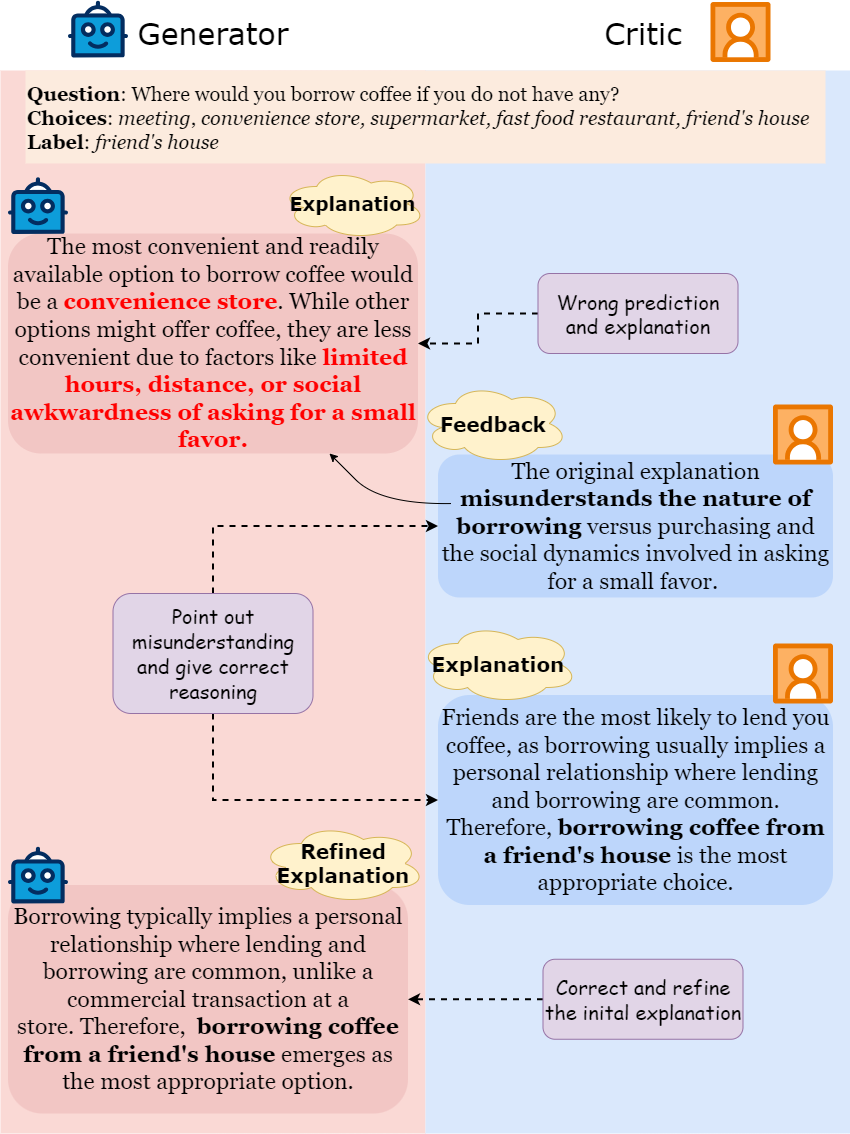}
\end{minipage}
}
\caption{\sys{Cross-Refine} example of the question \textit{``Where would you borrow coffee if you do not have any?''} from \data{ECQA}. The initial explanation by the \textbf{generator} has been accurately corrected and refined based on the feedback and explanations provided by the \textbf{critic}.}
\label{fig:dialogue_example}
\end{figure}

As the complexity of LLMs continues to increase, NLEs are pivotal in explainable artificial intelligence (XAI) \cite{madsen-2022-survey, lyu-2024-survey, zhao-2024-survey}. NLEs can serve as a bridge between XAI and humans, providing justifications in a format that humans can easily understand \cite{camburu-2018-esnli, wiegreffe-etal-2021-measuring}. LLMs are widely employed to generate NLEs across diverse domains \cite{singh-etal-2023-explaining, wang-2024-llmcheckup, kwon-2024-clinical, stern-2024-natural, wang-etal-2024-coxql}. However, similar to humans, LLMs may not consistently generate optimal explanations in their initial attempt \cite{madaan-2023-selfrefine}, e.g., due to lack of faithfulness \cite{chuang-2024-faithlm}. LLMs have the potential to enhance their reasoning abilities through self-improvement
without relying on external inputs \cite{huang-etal-2023-large}. Based on this observation, \citet{madaan-2023-selfrefine} proposed \sys{Self-Refine}, where LLMs use their own feedback to refine and improve their performance iteratively. This is shown to work only with large and powerful models; smaller models tend to hallucinate or generate repeated outputs. Moreover, \citet{tyen-etal-2024-llms} highlighted that LLMs generally struggle to identify reasoning errors and, therefore, cannot always self-correct their reasoning \cite{huang-2024-large}\looseness=-1. 

In this paper, we first propose \sys{Cross-Refine}, which draws inspiration from how humans benefit from learning from others \cite{foster-1995-learning, de-2023-learning} and additional feedback or suggestions. \sys{Cross-Refine} involves deploying a base LLM as the \textit{generator} to generate an NLE and a second LLM as the \textit{critic} (Figure~\ref{fig:dialogue_example}). While the generator outputs initial explanations, the critic provides the generator with feedback and suggestions based on initial explanations. Feedback and suggestions are then cross-referenced by the generator to refine the initial explanations. The cross-referencing process involves the refinement by the \textit{critic}, helping to mitigate the limitation of not being able to self-correct to some extent compared to \sys{Self-Refine} \cite{madaan-2023-selfrefine}. 

Secondly, we validate \sys{Cross-Refine} on three NLP tasks - commonsense question answering, natural language inference, and fact-checking. We perform an automatic evaluation using three model-based metrics, as well as a user study to assess explanations based on perceived faithfulness, insightfulness, and coherence. Both results suggest that \sys{Cross-Refine} can outperform \sys{Self-Refine} when LLMs have substantial knowledge relevant to the given task. However, when LLMs are required to reason about topics beyond their domain of expertise, e.g., in the medical domain, \sys{Cross-Refine} and \sys{Self-Refine} both perform poorly. We find that \sys{Cross-Refine} works effectively with less powerful LLMs, while \sys{Self-Refine} delivers strong results only with \lm{ChatGPT} \cite{OpenAI_ChatGPT_2023}.

Thirdly, compared to \sys{Self-Refine}, we incorporate the critic's feedback and suggestions instead of self-feedback. We conduct an ablation study to assess the importance of each deployed component. The ablation study reveals that both components contribute significantly and equally to the refinement of the explanations. 

Lastly, we evaluate \sys{Cross-Refine} on a bilingual dataset \data{HealthFC} \cite{vladika-etal-2024-healthfc} in English and German. The evaluation shows that \sys{Cross-Refine} can outperform \sys{Self-Refine} and consistently performs better in generating NLEs in German compared to \sys{Self-Refine}.

\begin{figure*}[ht!]
\centering
\resizebox{\textwidth}{!}{
\begin{minipage}{\textwidth}
\includegraphics[width=\linewidth]{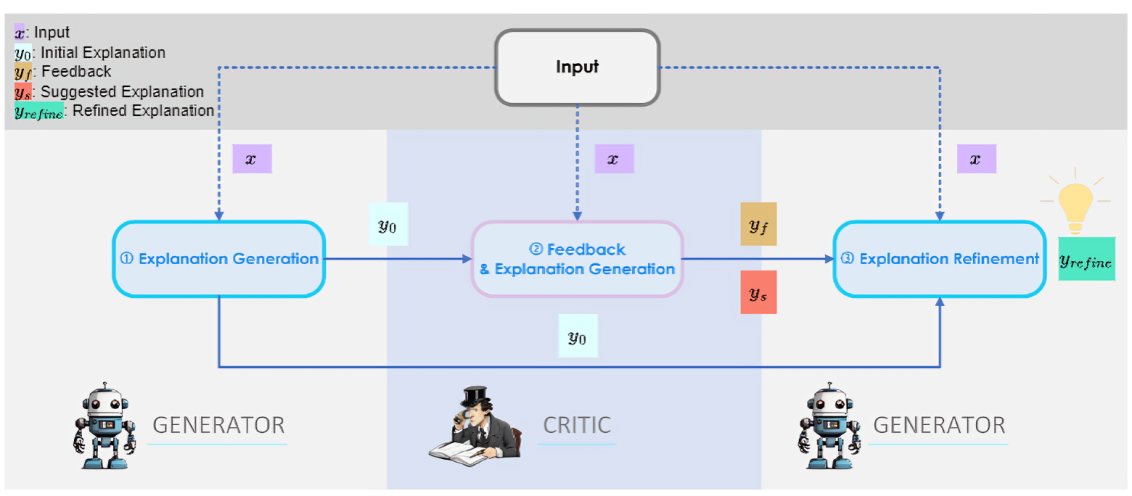}
\end{minipage}
}
\caption{Pipeline of \sys{Cross-Refine}. (1) Generator: produces an initial explanation.
(2): Critic: provides feedback and an suggested explanation based on the generator's initial output.
(3) Generator: utilizes the feedback and suggested explanation from the critic to refine and improve the initial explanation.}
\label{fig:pipeline}
\end{figure*}

\section{Background}
\label{sec:background}


\subsection{In-Context Learning for NLE}
Several Chain-of-Thought (CoT) \cite{wei-2022-cot} prompting techniques have been introduced that yield remarkable performance improvements in NLE generation, e.g., Zero-Shot CoT \cite{kojima-2022-zerocot}, Plan-and-Solve \cite{wang-etal-2023-plan}, and optimization by prompting \cite{yang-2024-large}. Self-consistency further demonstrates that self-evaluation can help LLMs improve reasoning \cite{wang-2023-selfconsistency}. \sys{Cross-Refine} also considers in-context learning to generate NLEs, and the critic deployed in \sys{Cross-Refine} plays a similar role to that of \citeposs{wang-2023-selfconsistency} self-evaluation. 

\subsection{NLE Evaluation} 
Regarding automated metrics for evaluating NLEs, BLEURT \cite{sellam-etal-2020-bleurt} calculates the semantic similarity between human annotated explanations and generated explanations. BARTScore \cite{yuan-2021-bartscore} treats the evaluation process as a text generation task and measures the likelihood of generating the reference text given the generated text. RORA \cite{jiang-2024-rora} measures the new information provided by a NLE to justify a label by evaluating the conditional $\nu$-information \cite{hewitt-etal-2021-conditional}. \citet{huang-etal-2024-chatgpt} asked \lm{ChatGPT} to evaluate the output of the generation on multiple scales. TIGERScore \cite{jiang-2024-tigerscore} uses natural language instructions to provide error analysis, pinpointing errors in the outputs.

For human evaluation of NLEs, prevalent metrics such as plausibility, faithfulness, simulatability, and insightfulness are used to evaluate factual correctness and logical coherence \cite{chan-etal-2022-comparative, atanasova-etal-2023-faithfulness}; consistency with the model's decision process \cite{lakkaraju-2019-faithfull, jacovi-goldberg-2020-towards, agarwal-2024-faithfulnessvsplausibilityunreliability}; how well a human can imitate model's behaviour based on explanations \cite{doshivelez-2017-rigorous, arora-2022-explain}; and how relevant is the
information of an explanation \cite{clinciu-etal-2021-study}, respectively. To validate \sys{Cross-Refine}, BLEURT, BARTScore, and TIGERScore are included for automatic evaluation (\S\ref{subsec:automatic_evaluation}), while perceived faithfulness, coherence, and insightfulness are included for the human evaluation (\S \ref{subsec:human_evaluataion}).


\section{Methodology}

\sys{Cross-Refine} is inspired by how humans learn from others and employs two LLMs separately for role modeling: one as the critic and the other as the generator (Figure~\ref{fig:pipeline}). The generator outputs the \textit{initial explanation}, while the critic offers \textit{feedback} and \textit{suggestions} on it, which can be used by the generator to refine the initial explanation.

\subsection{\sys{Cross-Refine} Example} 

Figure~\ref{fig:dialogue_example} provides an example for how generator and critic collaboratively improve NLEs. In the example, the generator initially chooses an incorrect choice (\textit{``convenience store''}), resulting in the explanation that is untruthful for the given question. In the feedback and suggested explanation provided by the critic, the errors made by the generator are identified (\textit{``misunderstanding the nature of borrowing''}), with the help of which the generator can recognize its mistakes and subsequently refine and correct both the prediction and the explanation\footnote{More \sys{Cross-Refine} examples are in Appendix~\ref{app:examples}.}.

\subsection{Pipeline}
We describe the pipeline of \sys{Cross-Refine} as shown in Figure~\ref{fig:pipeline} and denote the generator by $\mathcal{G}$ and the critic by $\mathcal{C}$.

\paragraph{Initial Generation}
The generator outputs the initial explanation independently using CoT prompting \cite{wei-2022-cot} with 3 to 20 shots depending on the input length (i.d. fewer shots with longer inputs) following the FEB template \cite{marasovic-etal-2022-shot}. Given an input \colorbox{purple}{$x$} and prompt $p_{gen}$, \sys{Cross-Refine} generates the initial explanation \colorbox{LightCyan}{$y_0$}:
\begin{equation}
    \colorbox{LightCyan}{$y_0$} = \mathcal{G}(\colorbox{purple}{$x$} | p_{gen})
\end{equation}


\paragraph{Quality Assessment}
Given an input \colorbox{purple}{$x$}, the initial explanation \colorbox{LightCyan}{$y_0$} and a prompt $p_{imp}$, the critic determines whether the initial explanation needs improvement \colorbox{Tan}{$y_{imp}$}: 

\begin{equation}
    \colorbox{Tan}{$y_{imp}$} = \mathcal{C}(\colorbox{LightCyan}{$y_0$}, \colorbox{purple}{$x$} | p_{imp})
\end{equation}

\paragraph{Feedback and Suggestion} Afterwards, the critic 
offers feedback \colorbox{orange}{$y_f$} on the initial explanation \colorbox{LightCyan}{$y_0$} from the generator based on the provided input \colorbox{purple}{$x$} with the prompt $p_f$:

\begin{equation}
     \colorbox{orange}{$y_f$} = \mathcal{C}(\colorbox{LightCyan}{$y_0$}, \colorbox{purple}{$x$} | p_f)
\end{equation}

Meanwhile, the critic generates a suggested explanation \colorbox{red}{$y_{s}$} by considering the input \colorbox{purple}{$x$}, the initial explanation \colorbox{LightCyan}{$y_0$}, and feedback \colorbox{orange}{$y_f$} generated by the critic with the prompt $p_s$:
\begin{equation}
    \colorbox{red}{$y_{s}$} = \mathcal{C} (\colorbox{orange}{$y_f$}, \colorbox{LightCyan}{$y_0$}, \colorbox{purple}{$x$} | p_s)
\end{equation}

\paragraph{Refinement}
Lastly, the feedback \colorbox{orange}{$y_f$} and the suggested explanation \colorbox{red}{$y_{s}$} generated by the critic are forwarded to the generator, which the generator uses to obtain the refined explanation \colorbox{gblue}{$y_{refine}$} with the prompt $p_{refine}$:
\begin{equation}
    \colorbox{gblue}{$y_{refine}$} = \mathcal{G} (\colorbox{red}{$y_{s}$}, \colorbox{orange}{$y_f$}, \colorbox{LightCyan}{$y_0$}, \colorbox{purple}{$x$} | p_{refine})
\end{equation}

In such a way, the generator can take into account the critic's \textit{feedback} and \textit{suggested explanation}. The feedback and suggested explanation are cross-referenced by the generator, which serves as a guide, ultimately enhancing the quality of the generator's initial explanations.

\section{Experimental Setup}
\label{sec:experiment}

\subsection{Baseline}
\label{subsec:baseline}
We employ \sys{Self-Refine} \cite{madaan-2023-selfrefine} as the baseline, which can enhance the initial outputs of the LLM only through iterative self-feedback. Unlike \sys{Cross-Refine}, it does not involve multiple reasoning steps and the model does not distinguish between the roles of critic and generator.

\subsection{Datasets}
\label{subsec:dataset}
Following \citet{atanasova-etal-2023-faithfulness}, we demonstrate the validity of our approach, \sys{Cross-Refine}, by applying it to three typical NLP tasks: natural language inference, commonsense question answering, and fact-checking. We select the subsequent three datasets\footnote{Examples from each dataset and label distributions of three employed dataset can be found in Appendix~\ref{app:dataset}.} because of their sufficient size and the high quality of human-annotated NLEs.

\paragraph{e-SNLI}
Natural Language Inference \cite{dagan-2006-nli} involves determining whether a given relationship between a \textit{premise} and a \textit{hypothesis} can be classified as \textit{entailment}, \textit{contradiction}, or \textit{neutrality}. The \data{e-SNLI} dataset \cite{camburu-2018-esnli} is an extension of the Stanford Natural Language Inference (SNLI) corpus \cite{bowman-etal-2015-large}, enriched with human-authored NLEs. 

\paragraph{ECQA} Compared to question answering, commonsense question answering requires the application of implicit background knowledge that extends beyond the information explicitly presented in the given context \cite{talmor-etal-2019-commonsenseqa}. Each instance in the \data{ECQA} dataset comprises a \textit{question}, several answer \textit{options}, and human annotated explanations \cite{aggarwal-etal-2021-explanations}.

\paragraph{HealthFC} The significance of fact-checking has greatly increased due to the swift spread of mis- and disinformation and  accurate information \cite{guo-etal-2022-survey}. \data{HealthFC} \cite{vladika-etal-2024-healthfc} is a bilingual fact-checking dataset (English and German) and consists of \textit{questions}, \textit{documents} as well as \textit{veracity annotations} (whether the answer is \textit{true}, \textit{false} or \textit{unknown} based on the provided document) and the corresponding \textit{explanations}. 

There are several reasons why we chose \data{HealthFC}\footnote{Note that due to the input length constraints, we extract only those sentences from the documents that were annotated as relevant in the original dataset. This results in shorter, more claim-focused documents that are then included in the prompt.}
 for our experiments. Firstly, this dataset is new and it is unlikely that it was seen during training by the employed LLMs. Secondly, it involves claims and documents from the medical domain and includes some specific terminology and domain knowledge that differs from more general-purpose data which LLMs are typically trained on. Thirdly, it is a bilingual dataset which means that we can check the performance of \sys{Cross-Refine} also with German.

\subsection{Models}
We select three state-of-the-art open-source general-purpose LLMs with increasing sizes from different model families: \lm{Qwen2-7B} \cite{yang-2024-qwen2technicalreport}, \lm{Mixtral-8x7B} \cite{jiang-2024-mixtral}, and \lm{Llama3-70B} \cite{llama3-2024-modelcard}\footnote{More details about models and inference time can be found in Appendix~\ref{app:experiment}.}.

\subsection{Demonstrations for In-Context Learning}
\label{subsec:dataset_creation}
To refine the initial explanation, we employ in-context learning to prompt the critic for feedback and suggestions and prompt the generator for refined explanations. For this purpose, we create a collection of demonstrations \data{FiXer}\footnote{Abbreviation of ``\underline{F}eedback of \underline{i}nitial e\underline{X}planation and \underline{e}xplanation \underline{r}efinement'' (\data{FiXer}). More details about data collection are provided in Appendix~\ref{app:demonstrations}.}, which comprises the initial explanations of the generator, the feedback and suggested explanations of the critic, and the refined explanations of the generator.

\subsection{Prompts}
\label{subsec:prompts}
Conforming to the FEB template \cite{marasovic-etal-2022-shot}, the prompt instructions used for explanation refinement include the task description, a list of information provided, and a few demonstrations for in-context learning (\S\ref{subsec:dataset_creation}), as depicted in Appendix~\ref{app:prompt_instruction}.

\section{Evaluation}
\label{sec:evaluation}

\subsection{Automatic Evaluation}
\label{subsec:automatic_evaluation}

The refined explanations are evaluated using the following three automated reference-based or reference-free metrics\footnote{The models used for automatic evaluation metrics are detailed in Appendix~\ref{app:automatic_evaluation_model}.}.



\paragraph{BLEURT} BLEURT \cite{sellam-etal-2020-bleurt} utilizes \lm{BERT} \cite{devlin-etal-2019-bert}, which is fine-tuned on a collection of human ratings, to deliver ratings of generated outputs, ranging from -1 to 1.

\paragraph{BARTScore} BARTScore \cite{yuan-2021-bartscore} leverages \lm{BART} \cite{lewis-etal-2020-bart} to score the generated text based on how well the generated text matches the reference text. Additionally, BARTScore evaluates both ``from generated to reference" and ``from reference to generated" directions, providing a more robust assessment.


\paragraph{TIGERScore} TIGERScore \cite{jiang-2024-tigerscore} utilizes natural language instructions to perform error analysis, identifying mistakes in the generated text using fine-tuned \lm{Llama2} \cite{touvron-2023-llama2openfoundation} and delivering corresponding explanations for each mistake. TIGERScore assigns a penalty score between $[-5, -0.5]$ for each mistake.


\subsection{Human Evaluation}
\label{subsec:human_evaluataion}
To further validate \sys{Cross-Refine}, we conduct a user study in which participants subjectively evaluate the refined explanations according to three dimensions. 

\subsubsection{Subjective Ratings}
\label{subsubsec:subjective_ratings}
Based on how \citet{feldhus-etal-2023-interrolang} and \citet{chiang-lee-2023-large} design Likert scales for explanation evaluation, we ask human annotators to assess reasoning outputs generated by \sys{Cross-Refine} based on the following dimensions used in the user study conducted by \citet{tsai-etal-2024-leveraging}:
\begin{itemize}[noitemsep,leftmargin=*]
    \item \textbf{Perceived Faithfulness (Binary)}: Investigate whether the generated reasoning exhibits hallucination and if it includes any misinformation;

    \item \textbf{Coherence (5-point Likert)}: Assess whether the generated reasoning is sensible, clear and coherent and reflects the reasons behind the user’s preference;
    
    \item \textbf{Insightfulness (5-point Likert)}: Evaluate the extent to which the generated reasoning provides informative insights into the user’s preferences.

\end{itemize} 
Coherence and insightfulness are rated on a 5-point Likert scale ranging from ``strongly disagree'' to ``strongly agree'', corresponding to points from 1 to 5. Perceived faithfulness is assessed using a binary scale, with a score of 0 assigned for unfaithful explanations and 1 for faithful explanations.


\subsubsection{User Study Setup}
Given the large number of combinations shown in Table~\ref{tab:automatic_evaluation}, we limit the user study to the easiest and most difficult datasets, \data{ECQA} and \data{HealthFC}, respectively. Additionally, we focus on \lm{Qwen2} and \lm{Llama3} as the generators, since \lm{Mixtral} does not perform well with \sys{Self-Refine} and \sys{Cross-Refine} (Table~\ref{tab:automatic_evaluation}). In this way, we maintain a feasible scope of our user study. 

We sample subsets ($n=10$) of \data{ECQA} and \data{HealthFC} randomly among the instances that fulfill the selection criteria described in Appendix \ref{app:sample_selection}, which makes the task more manageable for the annotators, reducing the risk of performance decline over time \cite{mangin-2022-plausible}, and ensuring the quality of the annotations. Based on the inputs, explanations are generated using different combinations of three deployed LLMs and the baseline, as illustrated in Table~\ref{tab:user_study}. Each explanation is rated by two annotators based on three subjective evaluation dimensions (\S\ref{subsubsec:subjective_ratings}). The inputs and their corresponding explanations are provided to the annotators in the form of questionnaires\footnote{The annotation instructions can be found in Appendix~\ref{app:annotation}.}. We use the Crowdee crowdsourcing platform\footnote{\url{https://www.crowdee.com/}} to recruit annotators, distribute questionnaires, and store the annotators' responses. We recruit a total of 32 annotators who are all English native speakers and do not necessarily have expertise in XAI.

\begin{table*}[bt!]
    \centering

    \resizebox{\textwidth}{!}{%
        \begin{tabular}{cc|ccc|ccc|ccc}

        \toprule
        \multicolumn{2}{c|}{Dataset} & \multicolumn{3}{c|}{\textbf{\data{ECQA}}} & \multicolumn{3}{c|}{\textbf{\data{eSNLI}}} & \multicolumn{3}{c}{\textbf{\data{HealthFC}}} \\
        \multirow{2}{*}{\textit{Critic}} & \multirow{2}{*}{\textit{Generator}} 
            & \multirow{2}{*}{BLEURT $\uparrow$} & BART \multirow{2}{*}{$\uparrow$} & TIGER \multirow{2}{*}{$\uparrow$}
            & \multirow{2}{*}{BLEURT $\uparrow$} & BART \multirow{2}{*}{$\uparrow$} & TIGER \multirow{2}{*}{$\uparrow$}
            & \multirow{2}{*}{BLEURT $\uparrow$} & BART \multirow{2}{*}{$\uparrow$} & TIGER \multirow{2}{*}{$\uparrow$}
            \\

        & & & Score \hspace{.5em} & Score \hspace{.5em} 
            & & Score \hspace{.5em} & Score \hspace{.5em} 
            & & Score \hspace{.5em} & Score \hspace{.5em} \\

        \midrule


        Self-Refine & \lm{Qwen2}
            & -0.68 & -3.91 & -4.38
            & -0.88 & -4.19 & -4.63
            & -0.25 & -3.09 & -1.09 \\ 
        
        \lm{Qwen2} & \lm{Qwen2} 
            & -0.33 & \textbf{-3.64} & -2.20
            & -0.97 & -3.33 & -4.33
            & \textbf{-0.24} & \textbf{-3.02} & \textbf{-0.79} \\ 

        \lm{Qwen2} & \lm{Mixtral} 
             & -0.67 & -4.13 & -2.88
            & \textbf{-0.71} & -3.44 & -3.65
            & -0.33 & -3.15 & -1.11 \\ 

        \lm{Qwen2} & \lm{Llama3} 
            & \textbf{-0.30} & -3.65 & \textbf{-1.71}
            & -0.99 & \textbf{-3.21} & \textbf{-2.55}
            & -0.83 & -3.60 & -2.87 \\ 

        \midrule


        Self-Refine & \lm{Mixtral}
            & -0.75 & -4.03 & -4.72
            & -0.83 & -3.72 & -4.50
            & -0.60 & -3.37 & -2.28 \\

        \lm{Mixtral} & \lm{Qwen2}
            & -0.50 & -4.08 & \textbf{-1.68}
            & -0.71 & \textbf{-3.44} & \textbf{-3.66}
            & -0.76 & -3.60 & -2.67 \\ 

        \lm{Mixtral} & \lm{Mixtral}
            & -0.66 & -3.98 & -2.25
            & \textbf{-0.64} & -3.49 & -3.87
            & \textbf{-0.38} & \textbf{-3.21} & \textbf{-1.41} \\ 

        \lm{Mixtral} & \lm{Llama3}
            & \textbf{-0.36} & \textbf{-3.46} & -4.48
            & -0.69 & -3.52 & -4.46
            & -0.81 & -3.61 & -2.87 \\ 

        \midrule


        Self-Refine & \lm{Llama3}
            & -0.59 & -3.79 & -5.64
            & -0.99 & -4.20 & -4.19
            & -0.33 & -3.14 & -1.85 \\ 
        
        \lm{Llama3} & \lm{Qwen2} 
             & \textbf{-0.37} & -3.72 & -2.72
            & \textbf{-0.51} & \textbf{-3.25} & -3.74
            & -0.76 & -3.55 & -2.65\\

        \lm{Llama3} & \lm{Mixtral} 
            & -0.45 & -3.64 & -3.78
            & -0.70 & -3.47 & \textbf{-3.43}
            & -0.30 & -3.13 & \textbf{-0.63}\\

        \lm{Llama3} & \lm{Llama3} 
            & -0.68 & \textbf{-3.62} & \textbf{-2.16}
            & -0.66 & -3.26 & -3.90
            & \textbf{-0.29} & \textbf{-3.07} & -0.97 \\
        
        \bottomrule
        \end{tabular}
    }
    \caption{Automatic evaluation results of refined explanations generated by \sys{Self-Refine}, and \sys{Cross-Refine} with \lm{Qwen2-7B}, \lm{Mixtral-8*7B}, and \lm{Llama3-70B} using BLEURT, BARTScore, and TIGERScore on the \data{ECQA}, \data{eSNLI}, and \data{HealthFC} datasets.
    }
    \label{tab:automatic_evaluation}
\end{table*}

\subsection{Ablation Study}
\label{subsec:ablation_study}
As illustrated in Figure~\ref{fig:pipeline}, the generator receives feedback and a suggested explanation from the critic in the final step to refine its initial explanation. To analyze the impact of individual components, namely \textit{feedback} and \textit{suggested explanation}, on the quality of the refined explanation, we conduct an comprehensive ablation experiment (\S \ref{subsec:eval_ablation_study}). 

\paragraph{Influence of Suggestions}
Compared to \sys{Self-Refine}, \sys{Cross-Refine} additionally introduces suggestions from the critic to guide the generator, we explore the extent to which the suggestions can influence the refined explanations.

\subsection{\sys{Cross-Refine} on German Data}
While the data we have described thus far is only in English, we also investigate the effectiveness of \sys{Cross-Refine} on the German data provided in \data{HealthFC} dataset (\S\ref{subsec:eval_german_data}).\looseness=-1 



\section{Results}
\subsection{Automatic Evaluation}
\label{subsec:res_automatic_evaluation}
Table~\ref{tab:automatic_evaluation} demonstrates that \sys{Cross-Refine} can easily outperform \sys{Self-Refine} on \data{ECQA} and \data{eSNLI}, although the scores for each automated metric are lower compared to the results of \data{HealthFC}. This discrepancy can be attributed to the shorter length of the gold rationales in \data{ECQA} and \data{eSNLI} relative to those in \data{HealthFC}. The longer context inherent in \sys{Cross-Refine}, which includes feedback and suggestions from the critic, tends to generate relatively longer explanations, contributing to this variation in scores.

Interestingly, Table~\ref{tab:automatic_evaluation} reveals that for \data{HealthFC}, \sys{Cross-Refine} with the same LLM as both generator and critic (``self \sys{Cross-Refine}'') outperforms \sys{Self-Refine}\footnote{``Self \sys{Cross-Refine}'' differs from \sys{Self-Refine} in that it additionally incorporates explicit suggestions from the critic. \sys{Self-Refine} instead just improves itself in few shots.}, indicating that suggestions play a crucial role in refining explanations (a further proof is shown in \S\ref{subsec:eval_ablation_study}). However, \sys{Cross-Refine} underperforms compared to \sys{Self-Refine} on \data{HealthFC} when using different combinations of LLMs instead of ``self \sys{Cross-Refine}''. The poorer performance might be caused by the lack of domain-specific knowledge, particularly in the medical domain, as the three LLMs that we deploy are general purpose models \cite{yang-2024-leveragingopenknowledgeadvancing}. Furthermore, since \data{HealthFC} was released very recently \cite{vladika-etal-2024-healthfc}, it is highly unlikely that three LLMs were trained on \data{HealthFC}, unlike the other two datasets. This result aligns with our intuition that models which lack knowledge in a particular domain are less likely to provide constructive and helpful feedback and suggestions to others \cite{valero-2019-effects}. Moreover, it suggests that cross-referencing could potentially lead to worse performance if feedback and suggestions are incorrect or hallucinated \cite{tan-2022-diversity, augenstein-2023-factualitychallengeseralarge}.

\subsection{User Study}
Table~\ref{tab:user_study} shows that, for \data{ECQA}, \sys{Cross-Refine} overall outperforms \sys{Self-Refine}, particularly in terms of coherence, where the margin is relatively large. Similarly, for \data{HealthFC}, the findings align with those mentioned in \S\ref{subsec:res_automatic_evaluation}: ``self \sys{Cross-Refine}'' can outperform \sys{Self-Refine}, but other combinations other than ``self \sys{Cross-Refine}'' perform worse than \sys{Self-Refine}. Furthermore, we discover a correlation between TIGERScore and the results of the user study.

Since each combination from Table~\ref{tab:user_study} is evaluated by two annotators, we report that our inter-annotator agreements (IAA) are at Krippendorff's $\alpha$ of 0.45 for \data{ECQA} and 0.39 for \data{HealthFC}. The low IAA scores can be attributed to the factor that we evaluate perceived faithfulness and insightfulness using a 5-point Likert scale, which is more fine-grained compared to a binary choice. The IAA on \data{HealthFC} is lower compared to \data{ECQA} due to its intrinsic difficulty. Additionally, we calculate the exact match between the two annotators, but in many cases, their scores are very close, such as 4 (agree) and 5 (strongly agree) or 1 (strongly disagree) and 2 (disagree).

From Table~\ref{tab:user_study}, we observe that the scores for perceived faithfulness are sightly higher for \data{HealthFC} compared to \data{ECQA}. In some cases, where medical domain knowledge is required, annotators might not fully grasp the context of instances from \data{HealthFC}, especially when the explanations seem to be plausible. Meanwhile, recruiting annotators with specific expertise, especially in the medical field, is very challenging through crowdsourcing platforms. Moreover, we lack the expertise in the medical domain to filter qualified recruited annotators. These findings can partially highlight the risks of over-trusting LLM outputs when individuals are not well-versed in the given topic \cite{li-etal-2024-think}. 

\begin{table}[t!]
    \centering
    \renewcommand*{\arraystretch}{}
    \footnotesize
    \resizebox{\columnwidth}{!}{%
        \begin{tabular}{cc|ccc|ccc}

        \toprule
        \multicolumn{2}{c|}{Dataset} & \multicolumn{3}{c|}{\textbf{\data{ECQA}}} & \multicolumn{3}{c}{\textbf{\data{HealthFC}}} \\
         \textit{Generator} & \textit{Critic} & Faith. & Coh. & Insight.  & Faith. & Coh. & Insight.  \\

        \midrule

        \multicolumn{2}{c|}{Self-Refine (\lm{Qwen})} 
            & 0.75 & 3.15 & 4.10
            & 0.75 & 3.75 & 3.90\\ 
        
        \lm{Qwen2} & \lm{Qwen2} 
            & \textbf{0.75} & \textbf{4.40 }& 4.05 
            & \textbf{1.00} & 3.20 & \textbf{4.15} \\ 

        \lm{Qwen2} & \lm{Mixtral} 
            & 0.50 & 3.80 & \textbf{4.15} 
            & 0.50 & 3.85 & 3.40 \\ 

        \lm{Qwen2} & \lm{Llama3} 
            & 0.50 & 3.65 & 3.20 
            & 0.25 & \textbf{4.20} & 3.80 \\ 

        \midrule


        \multicolumn{2}{c|}{Self-Refine (\lm{Llama3})} 
            & 0.50 & 2.80 & 2.35
            & 1.00 & 4.35 & 4.10\\ 
        
        \lm{Llama3} & \lm{Qwen2} 
            & 1.00 & 4.19 & 4.05 
            & 1.00 & 3.45 & 3.60  \\

        \lm{Llama3} & \lm{Mixtral} 
            & 0.75 & \textbf{4.50} & 4.00 
            & 0.50 & 3.15 & 2.75 \\

        \lm{Llama3} & \lm{Llama3} 
            & \textbf{1.00} & 4.05 &\textbf{ 4.15} 
            & \textbf{1.00} & \textbf{4.55} & \textbf{4.35}
            \\
        
        \bottomrule
        \end{tabular}
    }
    \caption{The results (in average scores from two annotators for each combination) of the user study on the quality of the refined explanations generated by \sys{Self-Refine} and \sys{Cross-Refine} using \lm{Qwen2} and \lm{Llama3} as the generator. The refined explanations are evaluated based on \textit{Perceived Faithfulness} (Faith.), \textit{Coherence} (Coh.), and \textit{Insightfulness} (Insight.), conducted on the \data{ECQA} and \data{HealthFC} datasets.
    }
    \label{tab:user_study}
\end{table}


Figure~\ref{fig:example_comparison} presents examples of \sys{Self-Refine} and \sys{Cross-Refine}. Like the examples shown in Figure~\ref{fig:example_comparison}, we observe several cases where the explanations generated by \sys{Self-Refine} are untruthful, while the critic in \sys{Cross-Refine} can correct errors, making the explanations more trustworthy. 

Therefore, based on automatic evaluation and the user study results, we can draw the conclusion that \sys{Cross-Refine} can outperform \sys{Self-Refine}, when LLMs possess substantial knowledge relevant to the given task. However, when LLMs are required to provide reasoning on topics outside of their domain of expertise, \sys{Cross-Refine} outperforms \sys{Self-Refine} only in the ``self \sys{Cross-Refine}'' setting, i.e. utilizing the same model for both critic and generator. 

\subsection{Ablation Study}
\label{subsec:eval_ablation_study}
For the evaluation, we randomly select samples from \data{eSNLI} and deploy \lm{Qwen2-7B} as both the generator and the critic, maintaining an analogous experimental setting to \sys{Self-Refine} \cite{madaan-2023-selfrefine}, as \sys{Self-Refine} shares the most similarity to our approach. We then generate explanations with and without a certain component and the automated metrics (\S \ref{subsec:automatic_evaluation}) are applied to each set of generated explanations to assess their quality.

\begin{table}[tb!]
    \centering
    \resizebox{\columnwidth}{!}{%
        \begin{tabular}{cc|ccc}

        \toprule
        \textbf{Feedback} & \textbf{Suggestion} & \textbf{BLEURT} & \textbf{BARTScore} & \textbf{TIGERScore} \\

        \midrule
        \checkmark & \checkmark & -0.97 & -3.33 & -4.33\\

        \checkmark & $\times$ & -0.72 \color{red}($\downarrow$ 0.25)  & -3.11 \color{green}($\uparrow$ 0.22) & -4.86 \color{red}($\downarrow$ 0.53) \\
        
        $\times$ & \checkmark & -0.84 \color{red}($\downarrow$ 0.13) & -3.35 \color{red}($\downarrow$ 0.02) & -4.67 \color{red}($\downarrow$ 0.34)\\
        
        
        \bottomrule
        \end{tabular}
    }
    \caption{
    Ablation Study of \sys{Cross-Refine}: Impact of different components on the refinement of explanations.
    }
    \label{tab:ablation_study}
\end{table}

Table~\ref{tab:ablation_study} shows that while BARTScore slightly increases when using \sys{Cross-Refine} without suggestions to refine the explanations, BLEURT and TIGERScore experience a sharp reduction. In contrast, when using \sys{Cross-Refine} without feedback, all scores decline to some extent, but not as significantly as in the case of \sys{Cross-Refine} without suggestions. Meanwhile, since we use the same LLM for both the generator and the critic in the ablation study (``self \sys{Cross-Refine}''), and \sys{Self-Refine} relies solely on self-feedback, making the feedback comparable between two approaches, we deduce that suggestions play an equally important role in the refinement of explanations.

\paragraph{Influence of Suggestions}
\label{subsec:eval_suggstion}

\begin{table}[t!]
    \centering
    \resizebox{\columnwidth}{!}{%
        \begin{tabular}{cc|cc|cc|cc}

        \toprule
        \multicolumn{2}{c|}{Model} & \multicolumn{2}{c|}{\textbf{\data{ECQA}}} & \multicolumn{2}{c|}{\textbf{\data{eSNLI}}} & \multicolumn{2}{c}{\textbf{\data{HealthFC}}} \\
        \textbf{Generator} & \textbf{Critic} & Init. & Sug. & Init. & Sug. & Init. & Sug.\\

        \midrule

        \lm{Qwen2} & \lm{Qwen2} & 0.76 & \textbf{0.87} & 0.45 & \textbf{0.85} & 0.90 & \textbf{0.96}\\ 

        \lm{Qwen2} & \lm{Mixtral} & \textbf{0.54} & 0.49 & 0.19 & \textbf{0.66} & 0.47 & \textbf{0.50} \\ 

        \lm{Qwen2} & \lm{Llama3} & 0.72 & \textbf{0.72} & 0.62 & \textbf{0.73} & 0.49 & \textbf{0.51}\\ 

        \midrule

        \lm{Mixtral} & \lm{Qwen2} & 0.46 & \textbf{0.50} & 0.37 & \textbf{0.84} & 0.56 & \textbf{0.78}\\ 

        \lm{Mixtral} & \lm{Mixtral} & 0.51 &\textbf{ 0.60} & 0.56 & \textbf{0.56} & 0.53 & \textbf{0.91}\\ 

        \lm{Mixtral} & \lm{Llama3} & 0.76 &\textbf{ 0.81} & 0.51 & \textbf{0.73} & 0.52 & \textbf{0.93}\\ 

        \midrule

        \lm{Llama3} & \lm{Qwen2} & 0.67 &\textbf{ 0.74} & 0.31 & \textbf{0.73} & 0.39 & \textbf{0.45} \\

        \lm{Llama3} & \lm{Mixtral} & \textbf{0.69} & 0.65 & 0.51 & \textbf{0.70} & 0.40 & \textbf{0.50}\\

        \lm{Llama3} & \lm{Llama3} & \textbf{0.63} & 0.61 & 0.46 & \textbf{0.64} & 0.73 & \textbf{0.92}\\

        \bottomrule
        \end{tabular}
    }
    \caption{
    The semantic similarities between the refined explanations and the initial explanations (\textbf{Init.}) and between the refined explanations and suggestions (\textbf{Sug.}).
    }
    \label{tab:cos_sim}
\end{table}
To measure the influence of suggestions, we evaluate the semantic similarity using \lm{SBERT}\footnote{\url{https://huggingface.co/sentence-transformers/all-mpnet-base-v2}} between the refined explanation and the initial explanation, as well as between the refined explanation and the suggestions individually. Table~\ref{tab:cos_sim} indicates that, in general, the refined explanations align more closely with the suggestions than with the initial explanations, which implies that the ``cross-refinement'' steps effectively prompt changes to the initial explanation. This process encourages LLMs to ``rethink'' and correct the initial explanations if they are stated incorrectly.

\subsection{\sys{Cross-Refine} on German Data}
\label{subsec:eval_german_data}
For automatic evaluation, we discard BLEURT and TIGERScore, as they only support English. For BARTScore, we use a different model that is compatible with German. In addition, we deploy MoverScore \cite{zhao-etal-2019-moverscore} and BERTScore \cite{zhang-2020-BERTScore}. MoverScore measures the semantic distance by contextualized representations and distance metrics, while BERTScore evaluates the token-level similarity between the reference texts and the LLM outputs by leveraging contextual embeddings. Table~\ref{tab:german} demonstrates that, overall, \sys{Cross-Refine} produces better NLEs than \sys{Self-Refine} on \data{HealthFC} (German). 

\begin{table}[bt!]
    \centering

    \resizebox{\columnwidth}{!}{%
        \begin{tabular}{cc|ccc}

        \toprule
        \multicolumn{2}{c|}{Dataset} &  \multicolumn{3}{c}{\textbf{\data{HealthFC} (German)}} \\
         \textit{Generator} & \textit{Critic} & BERTScore $\uparrow$ & BARTScore $\uparrow$ & MoverScore $\uparrow$  \\

        \midrule

        \multicolumn{2}{c|}{Self-Refine (\lm{Qwen})} & 0.6935
            & -5.6894  &  0.5246 \\ 
        
        \lm{Qwen2} & \lm{Qwen2} & \textbf{0.7068}
            & -4.4023  & \textbf{0.5271}  \\ 

        \lm{Qwen2} & \lm{Mixtral} & 0.6240
        & -6.5103 & 0.5068
              \\  

        \lm{Qwen2} & \lm{Llama3} & 0.7036
            & \textbf{-4.3785} & 0.5258 \\ 

        \midrule

        \multicolumn{2}{c|}{Self-Refine (\lm{Mixtral})} & 0.6519
            & -6.4200  & 0.5009 \\

        \lm{Mixtral} & \lm{Qwen2} & \textbf{0.6789}
           & -5.3713  & 0.5132 \\ 

        \lm{Mixtral} & \lm{Mixtral} & 0.6776
            & \textbf{-5.1785}  &  \textbf{0.5173}\\ 

        \lm{Mixtral} & \lm{Llama3}& 0.6782
            &-5.2327 &  0.5161\\ 

        \midrule

        \multicolumn{2}{c|}{Self-Refine (\lm{Llama3})} & 0.6626
            & -6.1267 &  0.5083 \\  
        
        \lm{Llama3} & \lm{Qwen2} & 0.6574
        & -5.6861   &  0.5078
              \\ 

        \lm{Llama3} & \lm{Mixtral} & 0.6220
        & -6.5031 &  0.5059
            \\ 

        \lm{Llama3} & \lm{Llama3} & \textbf{0.6656}
           & \textbf{-5.4474} &   \textbf{0.5088} \\ 
        
        \bottomrule
        \end{tabular}
    }
    \caption{Automatic evaluation results on \data{HealthFC} (German) dataset using BERTScore, BARTScore, and MoverScore. 
    }
    \label{tab:german}
\end{table}

For the German portion of the \data{HealthFC} dataset, we compare different configurations based on the number of explanations generated in German, English, or another language. Language identification is performed using \sys{fasttext-langdetect}\footnote{\url{https://github.com/zafercavdar/fasttext-langdetect}}. Table~\ref{tab:german-lang-statistics} presents the percentage of explanations generated in German. The results, summarized in Table \ref{tab:german-lang-statistics} show that \lm{Qwen2} and \lm{Llama3} consistently outperform \lm{Mixtral} in generating NLEs in German. Additionally, \sys{Self-Refine} outputs explanations in English notably more often compared to \sys{Cross-Refine}, e.g., \lm{Mixtral-8x7B} generates a higher percentage of self-refined explanations in English (57.5\%) compared to German (39\%), despite explicit prompts instructing that \textit{``Your response should be in German"} and several German-language demonstrations. 

Interestingly, in some cases, outputs are mixed, containing both English and German, e.g.: \textit{``Refined explanation: Die Antwort ist unbekannt, weil das Dokument aufzeigt ..."}, while English is typically used at the beginning to indicate the type of output, e.g., to indicate the type of the generated output such as \textit{``Refined explanation:"} in the example above. 

Overall, \sys{Cross-Refine} proves beneficial for generating explanations in a language different from English, even when the underlying model is predominantly trained on English data.
\begin{table}[bt!]
    \centering

    \resizebox{\columnwidth}{!}{%
        \begin{tabular}{cc|ccc}

        \toprule
        \multicolumn{2}{c|}{Dataset} &  \multicolumn{3}{c}{\textbf{\data{HealthFC} (German)}} \\
         \textit{Generator} & \textit{Critic} & German $\uparrow$ & English $\downarrow$ & Other $\downarrow$ \\

        \midrule

        \multicolumn{2}{c|}{Self-Refine (\lm{Qwen})} & 88.00 & 11.00 & 1.00 \\ 
        
        \lm{Qwen2} & \lm{Qwen2} & 96.86 & 2.29 & 0.86  \\ 

        \lm{Qwen2} & \lm{Mixtral} & 93.43 & 3.71 & 2.86
              \\  

        \lm{Qwen2} & \lm{Llama3} & \textbf{97.43} & \textbf{2.29} & \textbf{0.29} \\ 

        \midrule

        \multicolumn{2}{c|}{Self-Refine (\lm{Mixtral})} & 39.00 & 57.50 & 3.50 \\

        \lm{Mixtral} & \lm{Qwen2} & 74.29 & 23.43 & 2.29 \\ 

        \lm{Mixtral} & \lm{Mixtral} & \textbf{86.00} & \textbf{13.14} & \textbf{0.86} \\ 

        \lm{Mixtral} & \lm{Llama3}& 76.00 & 22.29 & 1.71 \\ 

        \midrule

        \multicolumn{2}{c|}{Self-Refine (\lm{Llama3})} & 57.50 & 41.00 & \textbf{1.50} \\  
        
        \lm{Llama3} & \lm{Qwen2} & 80.06 & 18.21 & 1.73 \\ 

        \lm{Llama3} & \lm{Mixtral} & \textbf{92.20} & \textbf{4.62} & 3.18 \\ 

        \lm{Llama3} & \lm{Llama3} & 82.95 & 14.74 & 2.31 \\ 
        
        \bottomrule
        \end{tabular}
    }
    \caption{Percentage of the generated explanations in different languages (English, German and other) for \data{HealthFC} (German).}
    \label{tab:german-lang-statistics}
\end{table}

\section{Related Work}
\label{sec:related_work}


\paragraph{Refined Explanations}
\citet{krishna-2023-amplify} proposed to take advantage of post-hoc explanations in in-context learning. \citet{tong-etal-2024-llms} found that LLMs can benefit from correct examples and learn from mistakes, while \citet{an-2024-learning} fine-tuned LLMs using pairs consisting of errors and their respective corrections. The mixture of agents (MoA) \cite{wang-2024-mixtureofagents} approach collects the strengths of multiple LLMs by constructing a layered MoA architecture and improves the reasoning by providing criticism. 
Moreover, LLMs can use self-generated feedback, refinement, or introspection as means to enhance reasoning abilities \cite{huang-etal-2023-large, madaan-2023-selfrefine, zhang-2023-selfconvinced, xu-etal-2024-sayself}. \citet{welleck-2023-generating} suggested to use a base generator that proposes an initial
hypothesis and a trained corrector that iteratively improves its quality. Compared to \citeposs{welleck-2023-generating} approach, \sys{Cross-Refine} does not necessarily require the critic can completely correct the hypothesis, as it can be very challenging \cite{huang-2024-large, tyen-etal-2024-llms}. Instead, \sys{Cross-Refine} focuses on providing feedback and suggested explanations generated by the critic, which the generator can use to refine its initial explanations. Furthermore, \sys{Cross-Refine} does not require supervised training data collection that is used for corrector fine-tuning \cite{welleck-2023-generating}. Meanwhile, \sys{Self-Refine} leverages self generated feedback to refine the explanation iteratively \cite{madaan-2023-selfrefine}. \citet{madaan-2023-selfrefine} showed that with \sys{Self-Refine}, less powerful LLMs struggle with explanation refinement, because they have difficulties in generating suitable feedback and thus tend to repeat the same output or generate hallucinated output. In comparison, since \sys{Cross-Refine} deploys the critic, the generator has an external source (i.e., feedback and suggestion) except for itself, which can be cross-referenced. Because of cross-reference, \sys{Cross-Refine} can be highly effective for tasks where LLMs have substantial knowledge. Moreover, \sys{Cross-Refine} performs well with less powerful LLMs, compared to \sys{Self-Refine}.





\section{Conclusion}
We introduced \sys{Cross-Refine}, an approach that improves NLEs through cross-refinement based on automated and human evaluation across various tasks. \sys{Cross-Refine} uses two LLMs for role modeling: One as the generator and the other as the critic. The generator refines its initial explanations by cross-referencing feedback and suggestions provided by the critic. Overall, \sys{Cross-Refine} can outperform similar state-of-the-art approaches such as \sys{Self-Refine} and can refine the explanations well with less powerful LLMs compared to \sys{Self-Refine}. For tasks that fall outside of the LLMs' domain expertise, e.g., in the medical domain, and require more structured domain knowledge, \sys{Cross-Refine} using the same LLM both as the generator and the critic can surpass \sys{Self-Refine}. Furthermore, since \sys{Cross-Refine} introduces feedback along with suggestions from the critic to refine the generator's initial explanation, through the ablation study, we observe that suggestions are as crucial as feedback in refining explanations. Additionally, we find that \sys{Cross-Refine} outperforms \sys{Self-Refine} when data is in German (\data{HealthFC}), and with \sys{Cross-Refine}, NLEs are more likely to be generated in German compared to \sys{Self-Refine}.


\section{Future Work}
Future work includes exploring whether human-crafted feedback and suggestions can align with LLM generated ones. We plan to conduct a more fine-grained error analysis to inspect to what extent \sys{Cross-Refine} can address the errors contained in the explanations. We will explore how the interpretation of terminology of quality metrics, e.g., faithfulness or insightfulness, can impact the quality of the user study. Furthermore, we will investigate whether \sys{Cross-Refine} using LLMs trained on medical data can perform better on the \data{HealthFC} dataset. In addition, we plan to incorporate human interactions into the \sys{Cross-Refine} workflow. 

\section*{Limitations}
\sys{Cross-Refine}, while not inherently iterative like \sys{Self-Refine}, already demonstrates superior performance compared to the latter. Moreover, its structure allows for straightforward adaptation into an adaptive framework, potentially enhancing its refinement capabilities further.

Despite we created a collection of demonstrations, \data{FiXer}, which includes various instances consisting of initial explanations (generator), feedback and suggested explanations (critic) and refined explanations (generator), we are limited to using a small number of demonstrations ($n \in [3, 10]$) depending on input length for few-shot prompting to refine NLEs with \sys{Cross-Refine}. This limitation is primarily due to constraints on context length, e.g. \lm{Mixtral 7*8B} has the context window with only 8k tokens\footnote{\url{https://huggingface.co/docs/transformers/en/model_doc/mixtral}}. We will consider using \sys{LongLLMLingua} \cite{jiang-etal-2024-longllmlingua} to compress the prompt while the model performance can be enhanced.

We only performed experiments using datasets in English and German (only for \data{HealthFC}). In other languages, current models might not offer the same advantages.

We have to use different automatic evaluation metrics or models for the German data in \data{HealthFC}, as BLEURT and TIGERScore do not support languages other than English.


\section*{Ethical statement} 
The conducted user study was ethically approved by the Ethics Committee of Faculty IV of Technische Universit\"at Berlin. The 32 annotators in our user study were paid at least the minimum wage according to the standards of our host institutions' regions. The annotation took each annotator 30 minutes on average.

\section*{Acknowledgment} 
We thank the anonymous reviewers of COLING 2025 for their helpful and rigorous feedback.
This work has been supported by the German Federal Ministry of Education and Research as part of the projects TRAILS (01IW24005) and VERANDA (16KIS2047).

\bibliography{custom}

\appendix

\section{\sys{Cross-Refine} Examples}
\label{app:examples}
\begin{figure}[ht!]
\centering
\resizebox{\columnwidth}{!}{
\begin{minipage}{\textwidth}
\includegraphics[width=\linewidth]{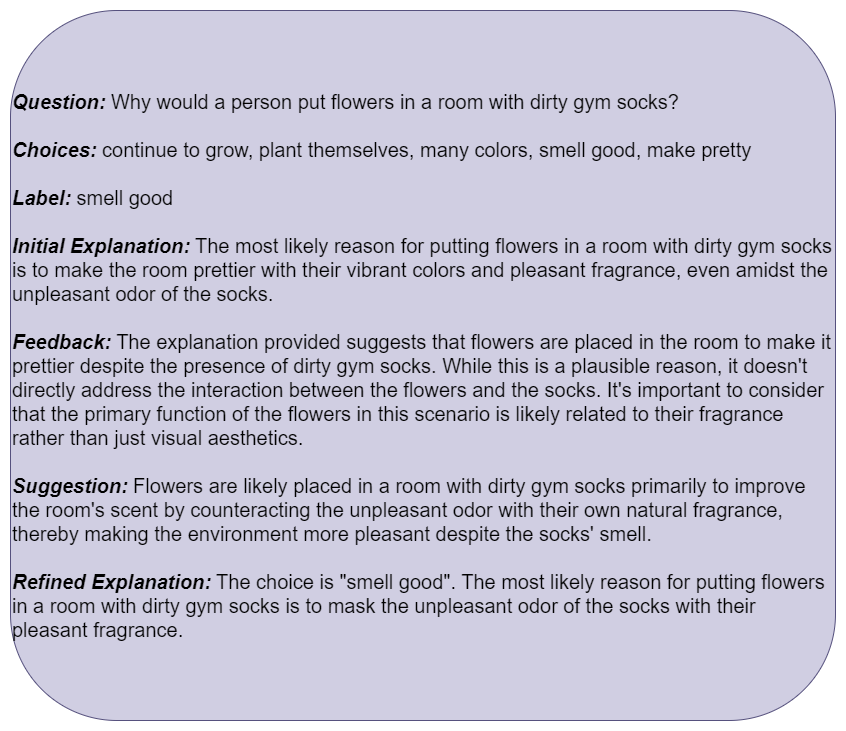}
\end{minipage}
}
\caption{\sys{Cross-Refine} example on \data{ECQA} dataset.}
\label{fig:ecqa}
\end{figure}

\begin{figure}[ht!]
\centering
\resizebox{\columnwidth}{!}{
\begin{minipage}{\textwidth}
\includegraphics[width=\linewidth]{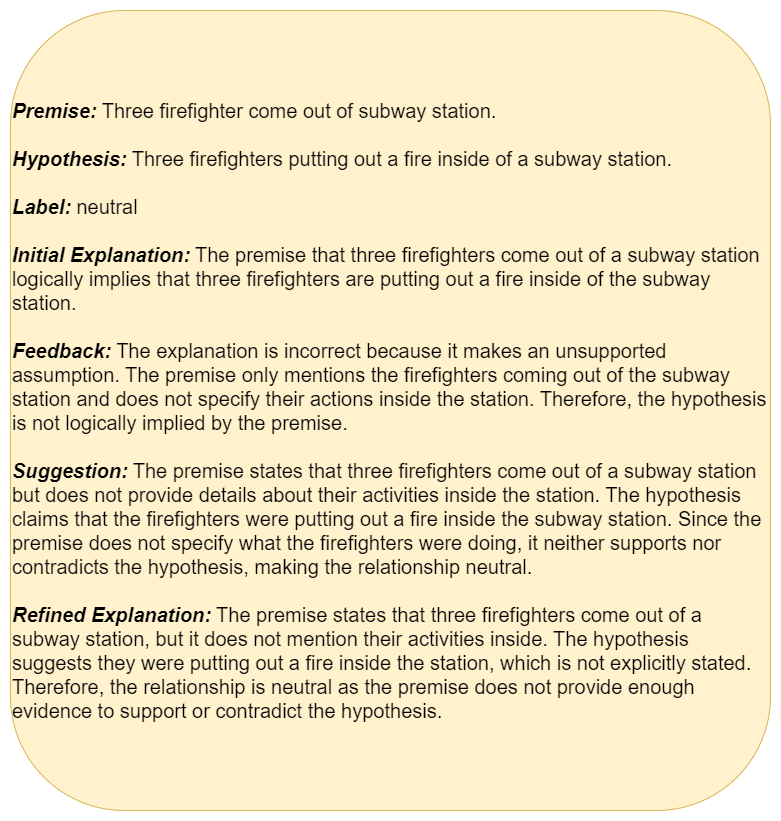}
\end{minipage}
}
\caption{\sys{Cross-Refine} example on \data{eSNLI} dataset.}
\label{fig:esnli}
\end{figure}

\begin{figure}[ht!]
\centering
\resizebox{\columnwidth}{!}{
\begin{minipage}{\textwidth}
\includegraphics[width=\linewidth]{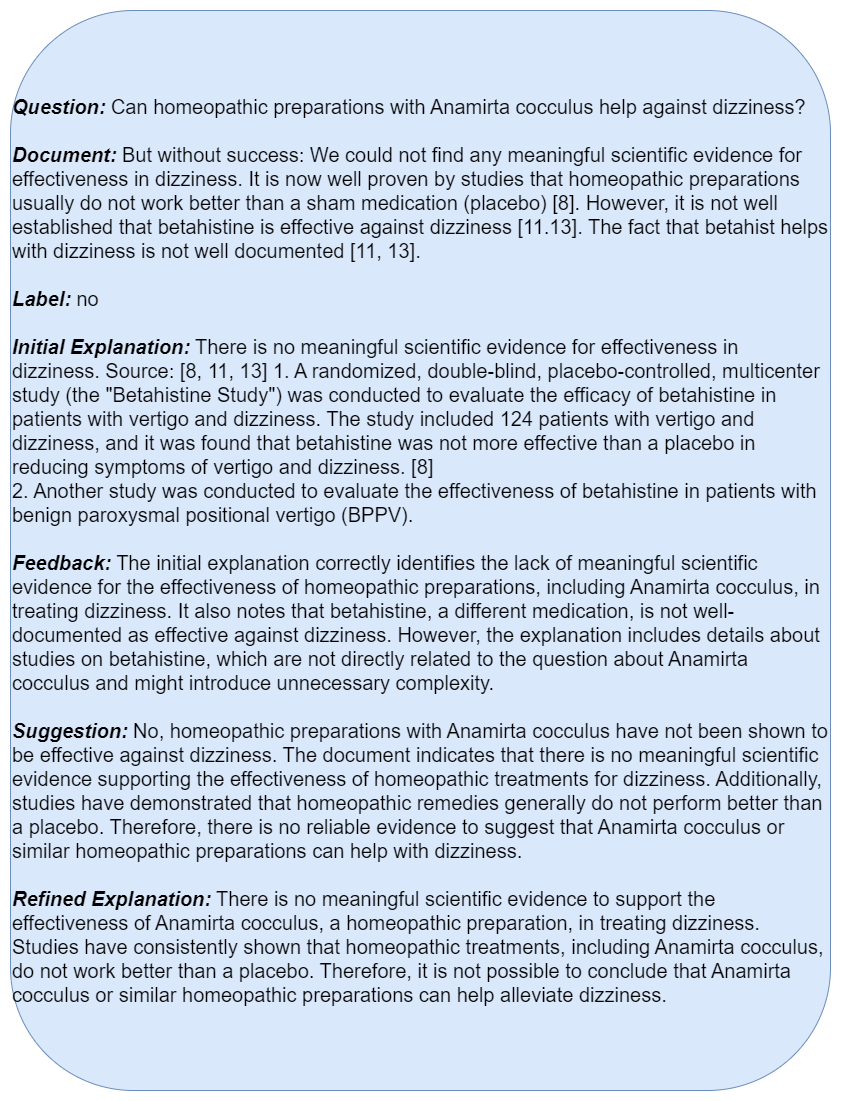}
\end{minipage}
}
\caption{\sys{Cross-Refine} example on \data{HealthFC} dataset.}
\label{fig:healthfc}
\end{figure}

Figure~\ref{fig:ecqa}, Figure~\ref{fig:esnli}, and Figure~\ref{fig:healthfc} show \sys{Cross-Refine} examples from \data{ECQA}, \data{eSNLI}, and \data{HealthFC}.

\section{Dataset}
\label{app:dataset}
\subsection{Dataset Example}
\label{app:dataset_example}
\begin{figure}[ht!]
\centering
\resizebox{\columnwidth}{!}{
\begin{minipage}{\textwidth}
\includegraphics[width=\linewidth]{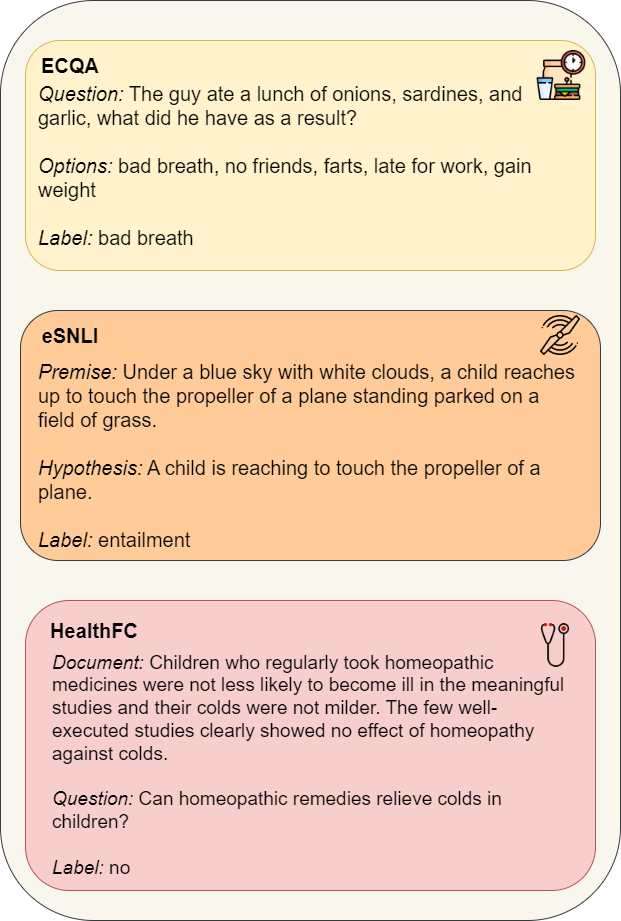}
\end{minipage}
}
\caption{Data points from \data{ECQA}, \data{eSNLI}, and \data{HealthFC}.}
\label{fig:dataset_example}
\end{figure}

Figure~\ref{fig:dataset_example} shows data points from \data{ECQA}, \data{eSNLI}, and \data{HealthFC}.

\subsection{Label Distribution}
\begin{figure}
  \centering

  \begin{subfigure}{\linewidth}
    \centering
    \centering
\includegraphics[width=\linewidth]{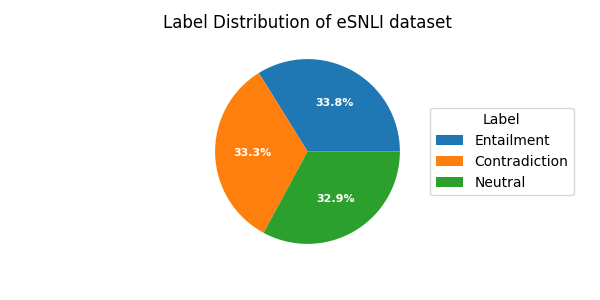}
\caption{\data{eSNLI} dataset}
\label{fig:esnli_label_distribution}
  \end{subfigure}
  \hfill
  \begin{subfigure}{\linewidth}
\centering
\includegraphics[width=\linewidth]{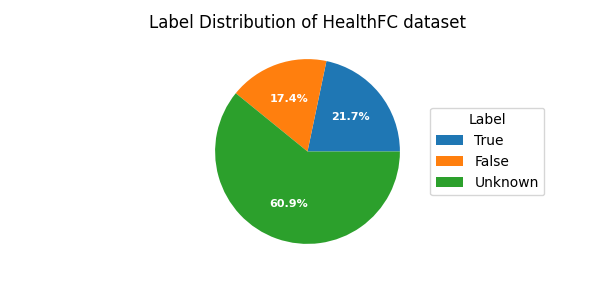}
\caption{\data{HealthFC} dataset}
\label{fig:healthfc_label_distribution}

  \end{subfigure}

  \caption{Label distributions of \data{eSNLI} and \data{HealthFC}.}
  \label{fig:label_distribution}
\end{figure}

Figure~\ref{fig:label_distribution} displays label distributions of \data{eSNLI} and \data{HealthFC}, as \data{ECQA} does not have fixed labels.

\section{Experiment}
\label{app:experiment}
\subsection{Models}
Table~\ref{tab:used_model} demonstrates LLMs that are used for \sys{Cross-Refine}. To reduce memory consumption, we use a GPTQ-quantized version \cite{frantar-2023-optq}. All LLMs are directly downloaded from Huggingface and run on a single NVIDIA RTXA6000, A100 or H100 GPU. 
\begin{table*}[t!]
    \centering
    \resizebox{\textwidth}{!}{%
        \begin{tabular}{rccc}

        \toprule
        \textbf{Name}& \textbf{Citation} & \textbf{Size} & \textbf{Link}\\

        \midrule
        \lm{Qwen2} & \cite{yang-2024-qwen2technicalreport} & 7B & \url{https://huggingface.co/Qwen/Qwen2-7B}\\
        
        \lm{Mixtral} & \citet{jiang-2024-mixtral} & 8*7B & \url{https://huggingface.co/mistralai/Mixtral-8x7B-v0.11} \\
        \lm{Llama3} & \cite{llama3-2024-modelcard} & 70B & \url{https://huggingface.co/meta-llama/Meta-Llama-3-70B}\\
        
        \bottomrule
        \end{tabular}
        }
    \caption{
    Three open sourced LLMs used in \sys{Cross-Refine}. 
    }
    \label{tab:used_model}
\end{table*}

\subsection{Inference Time}
\begin{table*}[t!]
    \centering
    \resizebox{\textwidth}{!}{%
        \begin{tabular}{c|cc|cc|cc}

        \toprule
         & \multicolumn{2}{c|}{\textbf{\data{ECQA}}} & \multicolumn{2}{c|}{\textbf{\data{eSNLI}}} & \multicolumn{2}{c}{\textbf{\data{HealthFC}}} \\
         Model & Feedback \& Suggestions & Refinement & Feedback \& Suggestions & Refinement & Feedback \& Suggestions & Refinement\\

        \midrule

        \lm{Qwen2-7B} & 2h & 5h & 2h & 4h & 6h & 14h \\

        \lm{Mixtral 8*7B} & 7h & 15h & 7h & 12h & 15h & 21h \\

        \lm{Llama3-70B} & 9h & 15h & 8h & 16h & 21h & 48h \\

        \bottomrule
        \end{tabular}
    }
    \caption{
    Inference time for feedback \& suggestions generation and refinement of explanations using \lm{Qwen2-7B}, \lm{Mixtral 8*7B} and \lm{Llama3-70B} on \data{ECQA}, \data{eSNLI} and \data{HealthFC}.
    }
    \label{tab:inference}
\end{table*}
Table~\ref{tab:inference} shows inference time for feedback \& suggestions generation and refinement of explanations using \lm{Qwen2-7B}, \lm{Mixtral 8*7B} and \lm{Llama3-70B} on \data{ECQA}, \data{eSNLI} and \data{HealthFC}.

\section{Demonstrations for In-Context Learning}
\label{app:demonstrations}
Firstly, we prompt \lm{Llama3-8B} \cite{llama3-2024-modelcard} to generate the initial explanations, which potentially has more room for improvement compared to larger LLMs\footnote{Note that for \data{HealthFC} (German), we use \lm{ChatGPT} instead of \lm{Llama3-8B} to ensure that the generated outputs are consistently in German.}.  Afterwards, we ask \lm{ChatGPT} to provide corresponding feedback and suggestions. Then we manually create a small subset of data points
that can be used as demonstrations for refining explanations, which are reviewed by two authors of this paper. Lastly, \lm{Llama3-8B} is prompted with created demonstrations to refine the initial explanations based on feedback and suggestions. The generated outputs then undergo a review process and are post-processed if necessary. For instance, if the initial explanation is of good quality and does not require improvement, or if the refined explanation is of lower quality than the initial explanation, we annotate whether examples need further refinement. Finally, we gather a total of 60 data points for \data{FiXer}. 

\section{Models Used for Automatic Evaluation Metrics}
\label{app:automatic_evaluation_model}
\begin{table*}[t!]
    \centering
    \resizebox{\textwidth}{!}{%
        \begin{tabular}{rcc}

        \toprule
        \textbf{Metric} & \textbf{Model} & \textbf{Link}\\

        \midrule
        BLEURT & \lm{BERT} & \url{https://huggingface.co/prajjwal1/bert-tiny}\\
        
        BARTScore & \lm{BART} & \url{https://huggingface.co/facebook/bart-large-cnn} \\
        TIGERScore & \lm{Llama2} & \url{https://huggingface.co/TIGER-Lab/TIGERScore-7B}\\
        BARTScore (DE) & \lm{mBART} & \url{https://huggingface.co/facebook/mbart-large-50}\\

        MoverScore & \lm{BERT} & \url{https://huggingface.co/google-bert/bert-base-german-cased} \\

         BERTScore & \lm{BERT} & \url{https://huggingface.co/google-bert/bert-base-german-cased} \\
        
        \bottomrule
        \end{tabular}
        }
    \caption{
    Models used for automatic evaluation
    metrics. 
    }
    \label{tab:automatic_evaluation_model}
\end{table*}
Table~\ref{tab:automatic_evaluation_model} displays the models used for automatic evaluation metrics.

\section{Prompt Instruction}
\label{app:prompt_instruction}

\begin{figure}[ht!]
\centering
\resizebox{\columnwidth}{!}{
\begin{minipage}{\columnwidth}
\includegraphics[width=\columnwidth]{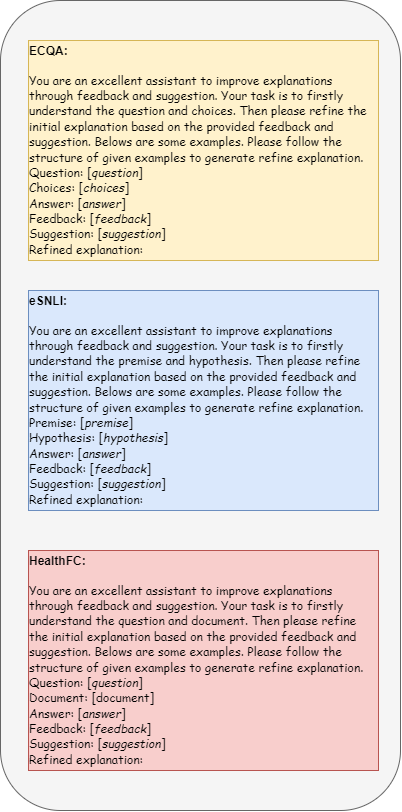}
\end{minipage}
}
\caption{Prompt instructions for \data{ECQA}, \data{eSNLI}, and \data{HealthFC}.}
\label{fig:prompt}
\end{figure}

The prompts used by \sys{Cross-Refine} for explanation refinement are given in Figure~\ref{fig:prompt}.

\section{User Study}
\label{app:annotation}

\begin{figure*}[ht!]
\centering
\resizebox{\textwidth}{!}{
\begin{minipage}{\textwidth}
\includegraphics[width=\linewidth]{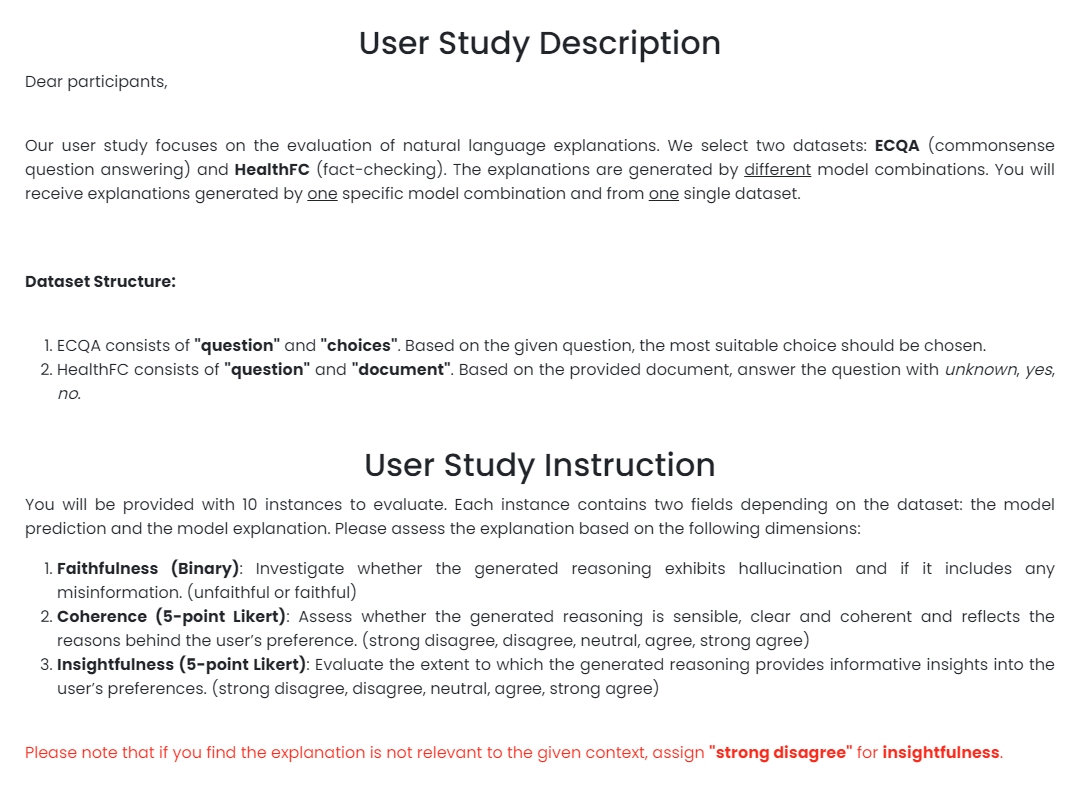}
\end{minipage}
}
\caption{Descriptions and instructions of the user study.}
\label{fig:user_study}
\end{figure*}

Figure~\ref{fig:user_study} displays the descriptions and instructions that we give the annotators for the user study.

\section{Sample Selection for User Study}
\label{app:sample_selection}

For the \data{HealthFC} dataset we observe different quality of generated explanations and to make sure that the explanations involved in the user study are meaningful we apply some selection criteria to filter out suboptimal generations (tokenization was performed with \sys{nltk}\footnote{\url{https://www.nltk.org/}} and cosine similarity was computed with \sys{SentenceTransformer}\footnote{\url{https://sbert.net/}} using the pre-trained model \lm{multi-qa-mpnet-base-cos-v1}\footnote{\url{https://huggingface.co/sentence-transformers/multi-qa-mpnet-base-cos-v1}}):

\begin{enumerate}
    \item Explanation length within 20 to 50 tokens.
    \item Bigram ratio: $\frac{num\_bigram\_types}{total\_num\_bigrams}>= 0.8$ to ensure the diversity of generated samples without too many repetitions of the same token(s).
    \item Digit ratio: $\frac{num\_digit\_tokens}{total\_num\_tokens}<= 0.3$ to ensure that the explanation does not contain too many digits.
    \item Cosine similarity between the embeddings of the original question and generated explanation is at least 0.6 to avoid including such cases where e.g. the model generates an explanation for one of the demonstrations instead of the input question-document pair.
\end{enumerate}

From those samples that fulfill all the requirements, we randomly sample 10 explanations per setting. The same procedure is applied to all combinations of models in both \sys{Self-Refine} and \sys{Cross-Refine} settings.

\section{Examples of \sys{Self-Refine} and \sys{Cross-Refine}}
\label{app:self_refine_cross_refine_examples}

\begin{figure*}
  \centering

  \begin{subfigure}{\linewidth}
    \centering
    \centering
\includegraphics[width=\columnwidth]{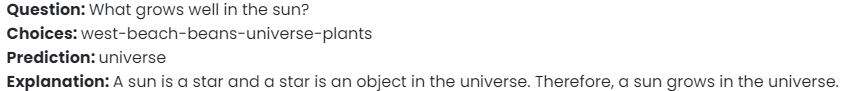}
\caption{\sys{Self-Refine}}
\label{fig:self-refine}
  \end{subfigure}
  \hfill
  \begin{subfigure}{\linewidth}
\centering
\includegraphics[width=\columnwidth]{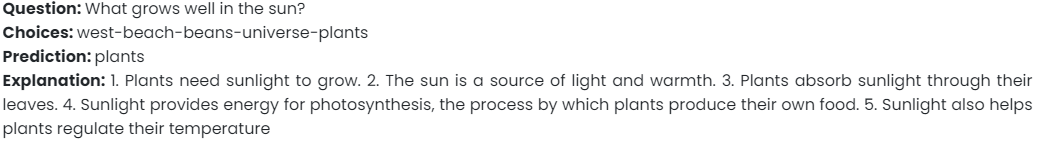}
\caption{\sys{Cross-Refine}}
\label{fig:cross-refine}

  \end{subfigure}

  \caption{Examples of \sys{Self-Refine} and \sys{Cross-Refine}. The gold label is \textit{plants}.}
  \label{fig:example_comparison}
\end{figure*}

Figure~\ref{fig:example_comparison} shows examples of \sys{Self-Refine} and \sys{Cross-Refine}.

\end{document}